\documentclass[10pt,twocolumn,letterpaper]{article}

\usepackage{cvpr}
\usepackage{times}
\usepackage{epsfig}
\usepackage{graphicx}
\usepackage{amsmath}
\usepackage{amssymb}
\usepackage{booktabs}
\usepackage{subcaption}
\usepackage{longtable}
\usepackage{mmstyle}

\usepackage{multirow}

\usepackage[pagebackref=true,breaklinks=true,letterpaper=true,colorlinks,bookmarks=false]{hyperref}

\newcommand{\codebase}{MMDetection}
\cvprfinalcopy 

\def\cvprPaperID{****} 

\begin{document}

\title{\codebase: Open MMLab Detection Toolbox and Benchmark}

\author{Kai Chen$^1$ \quad Jiaqi Wang$^{1}$\thanks{indicates equal contribution.} \quad Jiangmiao Pang$^{2*}$ \quad
Yuhang Cao$^1$ \quad Yu Xiong$^1$ \quad Xiaoxiao Li$^1$ \\
\quad Shuyang Sun$^3$ \quad Wansen Feng$^4$ \quad Ziwei Liu$^1$ \quad
Jiarui Xu$^5$ \quad Zheng Zhang$^6$ \quad Dazhi Cheng$^7$ \\
Chenchen Zhu$^8$ \quad Tianheng Cheng$^9$ \quad Qijie Zhao$^{10}$ \quad Buyu Li$^1$ \quad Xin Lu$^4$ \quad
Rui Zhu$^{11}$ \quad Yue Wu$^{12}$\\
Jifeng Dai$^6$ \quad Jingdong Wang$^6$ \quad Jianping Shi$^4$ \quad Wanli Ouyang$^3$ \quad Chen Change Loy$^{13}$ \quad Dahua Lin$^1$ \\ \\
{\small $^1$The Chinese University of Hong Kong \quad $^2$Zhejiang University \quad $^3$The University of Sydney \quad $^4$SenseTime Research} \\
{\small $^5$Hong Kong University of Science and Technology \quad $^6$Microsoft Research Asia \quad $^7$Beijing Institute of Technology} \\
{\small $^8$Nanjing University \quad $^9$Huazhong University of Science and Technology \quad $^{10}$Peking University} \\
{\small $^{11}$Sun Yat-sen University \quad $^{12}$Northeastern University \quad $^{13}$Nanyang Technological University} \\
}

\maketitle


\begin{abstract}

We present \codebase, an object detection toolbox that contains a rich set of
object detection and instance segmentation methods as well as related components and modules.
The toolbox started from a codebase of MMDet team who won the detection track of COCO Challenge 2018.
It gradually evolves into a unified platform that covers many popular detection methods and contemporary modules.
It not only includes training and inference codes, but also provides weights for more than 200 network models.
We believe this toolbox is by far the most complete detection toolbox.
In this paper, we introduce the various features of this toolbox.
In addition, we also conduct a benchmarking study on different methods,
components, and their hyper-parameters.
We wish that the toolbox and benchmark could serve the growing research
community by providing a flexible toolkit to reimplement existing methods and
develop their own new detectors.
Code and models are available at \url{https://github.com/open-mmlab/mmdetection}.
The project is under active development and we will keep this document updated.

\end{abstract}


\section{Introduction}
\label{sec:intro}

Object detection and instance segmentation are both fundamental computer
vision tasks. The pipeline of detection frameworks is usually more complicated
than classification-like tasks, and different implementation settings can lead
to very different results.
Towards the goal of providing a high-quality codebase and unified benchmark,
we build \codebase, an object detection and instance segmentation codebase
with PyTorch~\cite{paszke2017automatic}.

Major features of \codebase~are:
(1) \textbf{Modular design.}
We decompose the detection framework into different components and one can
easily construct a customized object detection framework by combining different
modules.
(2) \textbf{Support of multiple frameworks out of box.}
The toolbox supports popular and contempoary detection frameworks, see
Section~\ref{sec:supports} for the full list.
(3) \textbf{High efficiency.}
All basic bbox and mask operations run on GPUs. The training speed is faster
than or comparable to other codebases, including Detectron~\cite{Detectron2018}
, maskrcnn-benchmark~\cite{massa2018mrcnn} and SimpleDet~\cite{chen2019simpledet}.
(4) \textbf{State of the art.}
The toolbox stems from the codebase developed by the MMDet team, who won COCO
Detection Challenge in 2018, and we keep pushing it forward.

Apart from introducing the codebase and benchmarking results, we also report
our experience and best practice for training object detectors.
Ablation experiments on hyper-parameters, architectures, training strategies
are performed and discussed. We hope that the study can benefit future research
and facilitate comparisons between different methods.

The remaining sections are organized as follows. We first introduce various
supported methods and highlight important features of \codebase, and then present
the benchmark results. Lastly, we show some ablation studies on some chosen baselines.


\section{Supported Frameworks}
\label{sec:supports}

\codebase~contains high-quality implementations of popular object detection and
instance segmentation methods.
A summary of supported frameworks and features compared with other
codebases is provided in Table~\ref{tab:supported-features}. \codebase~supports
more methods and features than other codebases, especially for recent
ones. A list is given as follows.

\begin{table*}
	\centering
	\caption{Supported features of different codebases. ``\checkmark'' means officially supported, ``*'' means supported in a forked repository and blank means not supported.}
		\begin{tabular}{*{5}{c}}
			\toprule
			               & \codebase & maskrcnn-benchmark & Detectron & SimpleDet \\
			\midrule
            Fast R-CNN     & \checkmark & \checkmark & \checkmark & \checkmark \\
            Faster R-CNN   & \checkmark & \checkmark & \checkmark & \checkmark \\
            Mask R-CNN     & \checkmark & \checkmark & \checkmark & \checkmark \\
            RetinaNet      & \checkmark & \checkmark & \checkmark & \checkmark \\
            DCN            & \checkmark & \checkmark & \checkmark & \checkmark \\
            DCNv2          & \checkmark & \checkmark &            &            \\
            Mixed Precision Training & \checkmark & \checkmark &            & \checkmark \\
            Cascade R-CNN  & \checkmark &            & *          & \checkmark \\
            Weight Standardization & \checkmark & *  &            &            \\
            Mask Scoring R-CNN & \checkmark & *      &            &            \\
            FCOS           & \checkmark & *          &            &            \\
            SSD            & \checkmark &            &            &            \\
            R-FCN          & \checkmark &            &            &            \\
            M2Det          & \checkmark &            &            &            \\
            GHM            & \checkmark &            &            &            \\
            ScratchDet     & \checkmark &            &            &            \\
            Double-Head R-CNN & \checkmark &         &            &            \\
            Grid R-CNN     & \checkmark &            &            &            \\
            FSAF           & \checkmark &            &            &            \\
            Hybrid Task Cascade & \checkmark &       &            &            \\
            Guided Anchoring & \checkmark &          &            &            \\
            Libra R-CNN    & \checkmark &            &            &            \\
            Generalized Attention & \checkmark &     &            &            \\
            GCNet          & \checkmark &            &            &            \\
            HRNet          & \checkmark &            &            &            \\
            TridentNet~\cite{li2019scaleaware} &  &  &            & \checkmark \\
			\bottomrule
		\end{tabular}
	\label{tab:supported-features}
\end{table*}

\subsection{Single-stage Methods}

\begin{itemize}
\item SSD~\cite{liu2016ssd}: a classic and widely used single-stage detector with simple model architecture, proposed in 2015.
\item RetinaNet~\cite{lin2017focal}: a high-performance single-stage detector with Focal Loss, proposed in 2017.
\item GHM~\cite{li2019gradient}: a gradient harmonizing mechanism to improve single-stage detectors, proposed in 2019.
\item FCOS~\cite{tian2019fcos}: a fully convolutional anchor-free single-stage detector, proposed in 2019.
\item FSAF~\cite{zhu2019feature}: a feature selective anchor-free module for single-stage detectors, proposed in 2019.
\end{itemize}

\subsection{Two-stage Methods}

\begin{itemize}
	\item Fast R-CNN~\cite{girshick2015fast}: a classic object detector which requires pre-computed proposals, proposed in 2015.
	\item Faster R-CNN~\cite{ren2015faster}: a classic and widely used two-stage object detector which can be trained end-to-end, proposed in 2015.
	\item R-FCN~\cite{dai2016r}: a fully convolutional object detector with faster speed than Faster R-CNN, proposed in 2016.
	\item Mask R-CNN~\cite{he2017mask}: a classic and widely used object detection and instance segmentation method, proposed in 2017.
	\item Grid R-CNN~\cite{lu2019grid}: a grid guided localization mechanism as an alternative to bounding box regression, proposed in 2018.
	\item Mask Scoring R-CNN~\cite{huang2019mask}: an improvement over Mask R-CNN by predicting the mask IoU, proposed in 2019.
	\item Double-Head R-CNN~\cite{wu2019rethinking}: different heads for classification and localization, proposed in 2019.
\end{itemize}

\subsection{Multi-stage Methods}

\begin{itemize}
\item Cascade R-CNN~\cite{cai2018cascade}: a powerful multi-stage object detection method, proposed in 2017.
\item Hybrid Task Cascade~\cite{chen2019hybrid}: a multi-stage multi-branch object detection and instance segmentation method, proposed in 2019.
\end{itemize}

\subsection{General Modules and Methods}

\begin{itemize}
    \item Mixed Precision Training~\cite{micikevicius2018mixed}: train deep neural networks using half precision floating point (FP16) numbers, proposed in 2018.
    \item Soft NMS~\cite{bodla2017soft}: an alternative to NMS, proposed in 2017.
    \item OHEM~\cite{shrivastava2016training}: an online sampling method that mines hard samples for training, proposed in 2016.
    \item DCN~\cite{dai2017deformable}: deformable convolution and deformable RoI pooling, proposed in 2017.
    \item DCNv2~\cite{zhu2019deformable}: modulated deformable operators, proposed in 2018.
    \item Train from Scratch~\cite{he2018rethinking}: training from random initialization instead of ImageNet pretraining, proposed in 2018.
    \item ScratchDet~\cite{zhu2018scratchdet}: another exploration on training from scratch, proposed in 2018.
    \item M2Det~\cite{zhao2018m2det}: a new feature pyramid network to construct more effective feature pyramids, proposed in 2018.
    \item GCNet~\cite{cao2019GCNet}: global context block that can efficiently model the global context, proposed in 2019.
    \item Generalized Attention~\cite{zhu2019empirical}: a generalized attention formulation, proposed in 2019.
    \item SyncBN~\cite{Peng2018megdet}: synchronized batch normalization across GPUs, we adopt the official implementation by PyTorch.
    \item Group Normalization~\cite{wu2018group}: a simple alternative to BN, proposed in 2018.
    \item Weight Standardization~\cite{weightstandardization}: standardizing the weights in the convolutional layers for micro-batch training, proposed in 2019.
    \item HRNet~\cite{SunXLW19, sun2019high}: a new backbone with a focus on learning reliable high-resolution representations, proposed in 2019.
    \item Guided Anchoring~\cite{wang2019region}: a new anchoring scheme that predicts sparse and arbitrary-shaped anchors, proposed in 2019.
	\item Libra R-CNN~\cite{pang2019libra}: a new framework towards balanced learning for object detection, proposed in 2019.
\end{itemize}


\section{Architecture}
\label{sec:architecture}

\subsection{Model Representation}

Although the model architectures of different detectors are different, they
have common components, which can be roughly summarized into the following
classes.

\noindent\textbf{Backbone}
Backbone is the part that transforms an image to feature maps, such as a
ResNet-50 without the last fully connected layer.

\noindent\textbf{Neck}
Neck is the part that connects the backbone and heads. It performs some
refinements or reconfigurations on the raw feature maps produced by the backbone.
An example is Feature Pyramid Network (FPN).

\noindent\textbf{DenseHead (AnchorHead/AnchorFreeHead)}
DenseHead is the part that operates on dense locations of feature maps,
including AnchorHead and AnchorFreeHead, \eg, RPNHead, RetinaHead, FCOSHead.

\noindent\textbf{RoIExtractor}
RoIExtractor is the part that extracts RoI-wise features from a single or
multiple feature maps with RoIPooling-like operators. An example that extracts
RoI features from the corresponding level of feature pyramids is SingleRoIExtractor.

\noindent\textbf{RoIHead (BBoxHead/MaskHead)}
RoIHead is the part that takes RoI features as input and make RoI-wise task-specific
predictions, such as bounding box classification/regression, mask prediction.

\begin{figure}
    \centering
    \includegraphics[width=\linewidth]{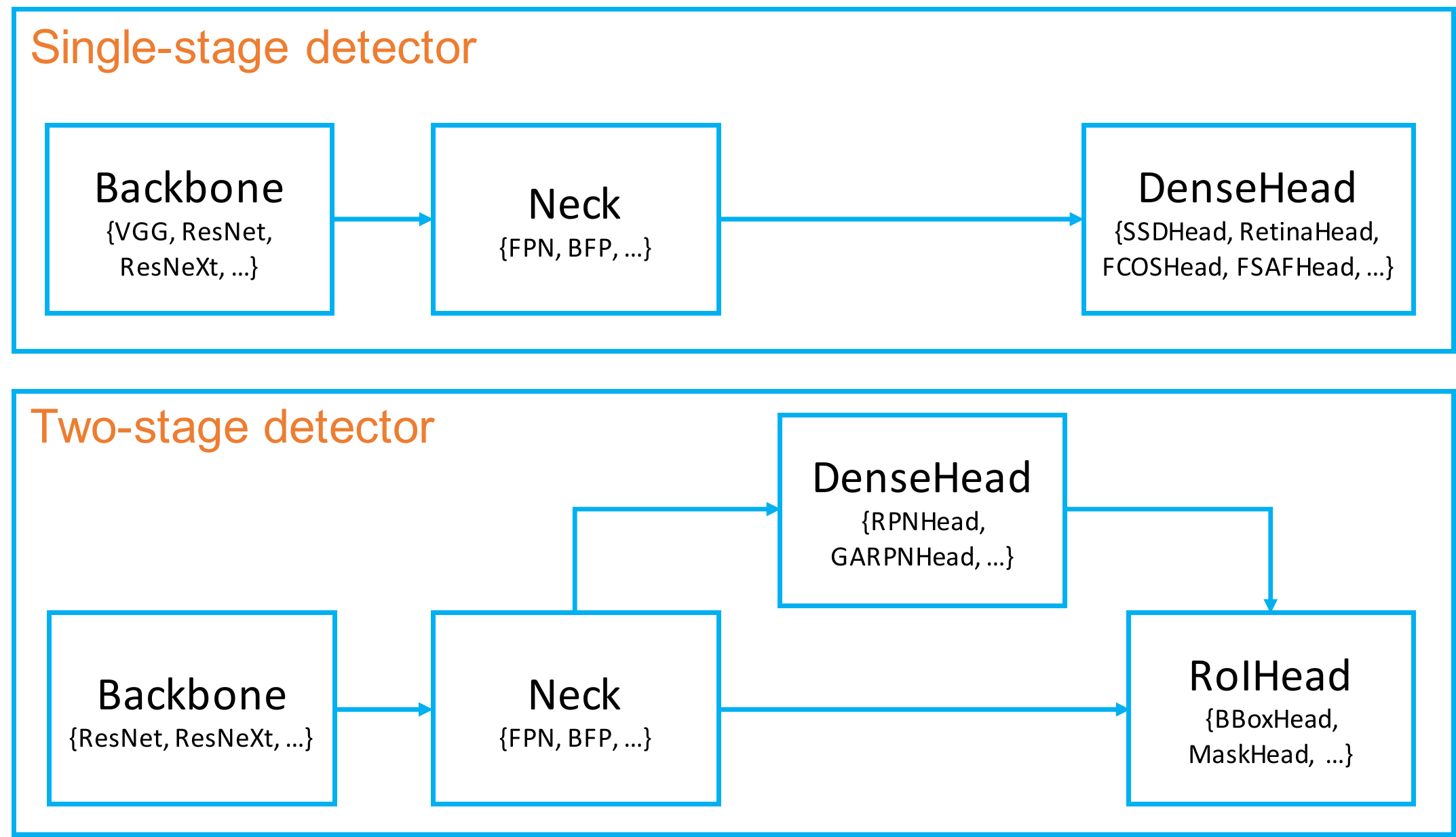}
	\caption{Framework of single-stage and two-stage detectors, illustrated with abstractions in \codebase.}
    \label{fig:framework}
\end{figure}

With the above abstractions, the framework of single-stage and two-stage
detectors is illustrated in Figure~\ref{fig:framework}.
We can develop our own methods by simply creating some new components and
assembling existing ones.

\subsection{Training Pipeline}

We design a unified training pipeline with hooking mechanism. This training
pipeline can not only be used for object detection, but also other computer
vision tasks such as image classification and semantic segmentation.

\begin{figure}
    \centering
    \includegraphics[width=\linewidth]{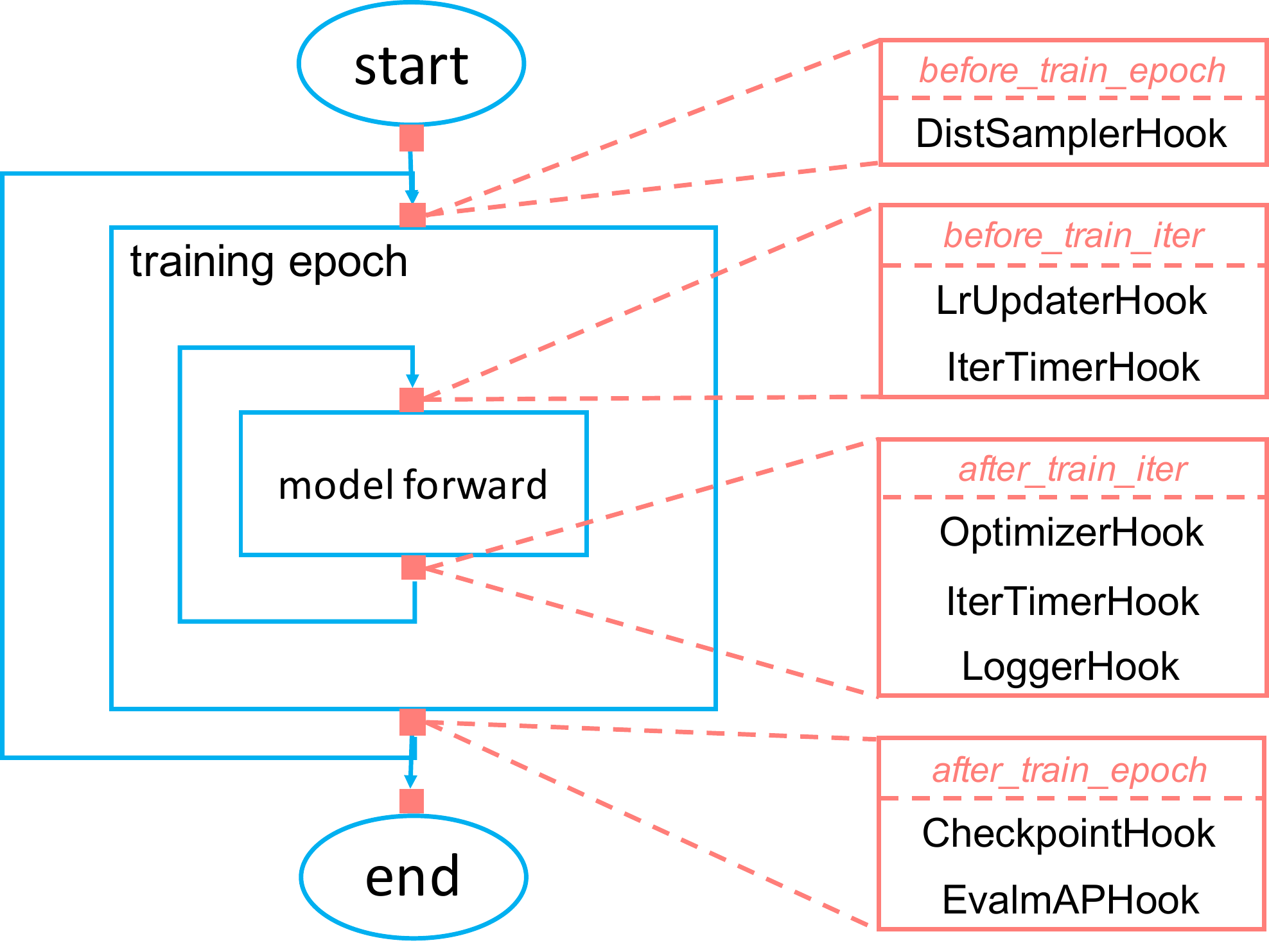}
	\caption{Training pipeline.}
    \label{fig:pipeline}
\end{figure}

The training processes of many tasks share a similar workflow, where training
epochs and validation epochs run iteratively and validation epochs are optional.
In each epoch, we forward and backward the model by many iterations.
To make the pipeline more flexible and easy to customize, we define a minimum
pipeline which just forwards the model repeatedly. Other behaviors are defined
by a hooking mechanism.
In order to run a custom training process, we may want to perform some
self-defined operations before or after some specific steps.
We define some timepoints where users may register any executable methods (hooks),
including \emph{before\_run, before\_train\_epoch, after\_train\_epoch,
before\_train\_iter, after\_train\_iter, before\_val\_epoch, after\_val\_epoch,
before\_val\_iter, after\_val\_iter, after\_run}.
Registered hooks are triggered at specified timepoints following the priority
level. A typical training pipeline in \codebase~is shown in Figure~\ref{fig:pipeline}.
The validation epoch is not shown in the figure since we use evaluation hooks
to test the performance after each epoch. If specified, it has the same
pipeline as the training epoch.

\section{Benchmarks}
\label{sec:experiments}

\subsection{Experimental Setting}

\noindent\textbf{Dataset.}
\codebase~supports both VOC-style and COCO-style datasets.
We adopt MS COCO 2017 as the primary benchmark for all experiments since it is more challenging and widely used.
We use the \emph{train} split for training and report the performance
on the \emph{val} split.

\noindent\textbf{Implementation details.}
If not otherwise specified, we adopt the following settings.
(1) Images are resized to a maximum scale of $1333\times800$,without changing the aspect ratio.
(2) We use 8 V100 GPUs for training with a total batch size of 16 (2 images per GPU) and a single V100 GPU for inference.
(3) The training schedule is the same as Detectron~\cite{Detectron2018}. ``1x'' and ``2x''
means 12 epochs and 24 epochs respectively. ``20e'' is adopted in cascade models, which denotes 20 epochs.

\noindent\textbf{Evaluation metrics.}
We adopt standard evaluation metrics for COCO dataset, where multiple IoU
thresholds from 0.5 to 0.95 are applied.
The results of region proposal network (RPN) are measured with Average Recall (AR)
and detection results are evaluated with mAP.

\subsection{Benchmarking Results}

\begin{figure*}[hbt]
	\centering
	\begin{subfigure}{.48\linewidth}
		\centering
    	\includegraphics[width=\linewidth]{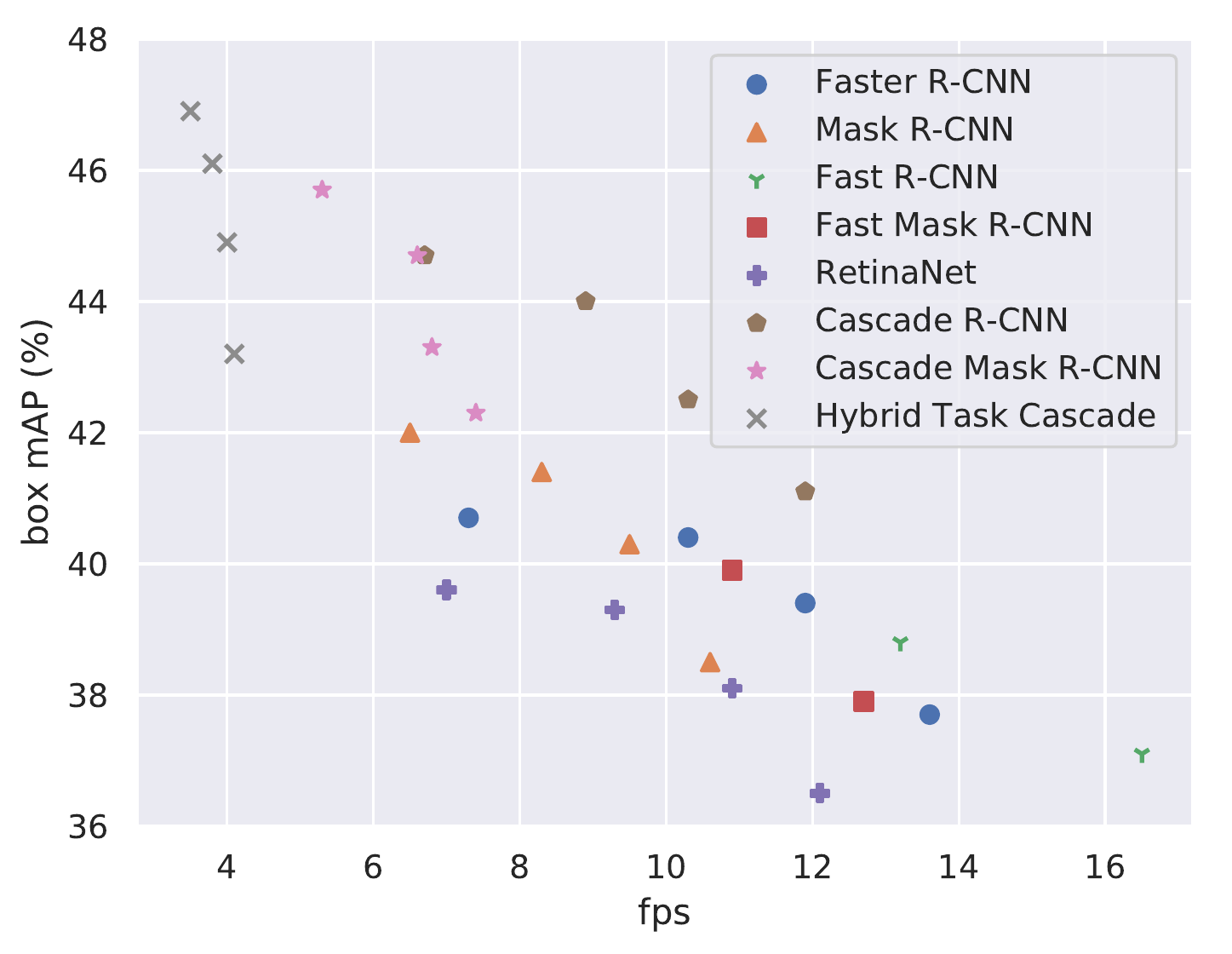}
	\end{subfigure}
	\begin{subfigure}{.48\linewidth}
		\centering
    	\includegraphics[width=\linewidth]{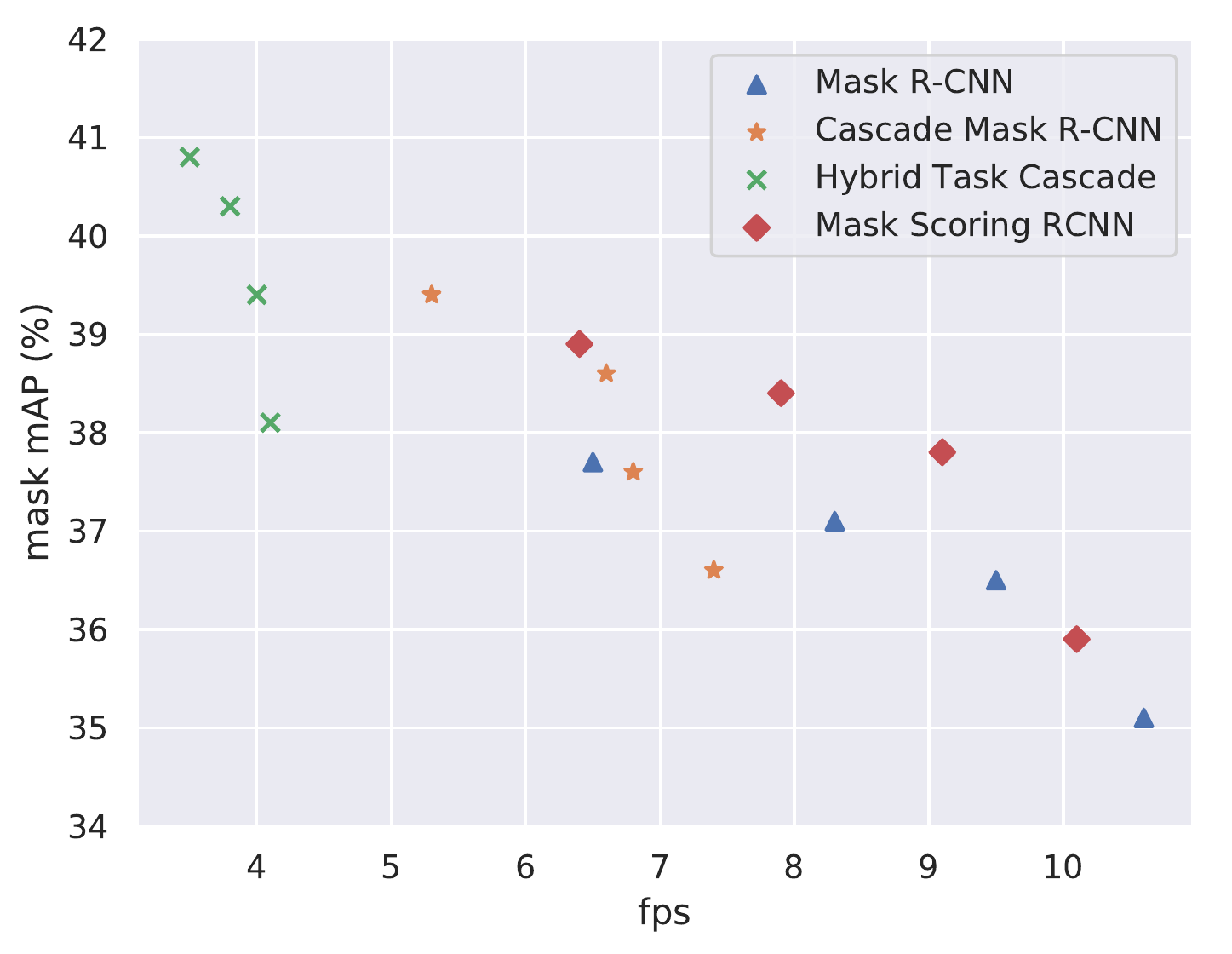}
	\end{subfigure}
	\caption{Benchmarking results of different methods. Each method is tested with four different backbones.}
	\label{fig:benchmark}
\end{figure*}

\noindent\textbf{Main results.}
We benchmark different methods on COCO 2017 \emph{val}, including SSD~\cite{liu2016ssd}, RetinaNet~\cite{lin2017focal},
Faster RCNN~\cite{ren2015faster}, Mask RCNN~\cite{he2017mask} and Cascade R-CNN~\cite{lin2017focal},
Hybrid Task Cascade~\cite{chen2019hybrid} and FCOS~\cite{tian2019fcos}.
We evalute all results with four widely used backbones, \ie,
ResNet-50~\cite{he2016deep}, ResNet-101~\cite{he2016deep},
ResNet-101-32x4d~\cite{xie2017aggregated} and ResNeXt-101-64x4d~\cite{xie2017aggregated}.
We report the inference speed of these methods and bbox/mask AP in Figure~\ref{fig:benchmark}.
The inference time is tested on a single Tesla V100 GPU.

\noindent
\textbf{Comparison with other codebases}
Besides \codebase, there are also other popular codebases like Detectron~\cite{Detectron2018},
maskrcnn-benchmark~\cite{massa2018mrcnn} and SimpleDet~\cite{chen2019simpledet}.
They are built on the deep learning frameworks of caffe2\footnote{\url{https://github.com/facebookarchive/caffe2}},
PyTorch~\cite{paszke2017automatic} and MXNet~\cite{chen2015mxnet}, respectively.
We compare \codebase~with Detectron (@a6a835f), maskrcnn-benchmark (@c8eff2c)
and SimpleDet (@cf4fce4) from three aspects: performance, speed and memory.
Mask R-CNN and RetinaNet are taken for representatives of two-stage and
single-stage detectors.
Since these codebases are also under development, the reported results in their
model zoo may be outdated, and those results are tested on different hardwares.
For fair comparison, we pull the latest codes and test them in the same
environment.
Results are shown in Table~\ref{tab:codebase-compare}.
The memory reported by different frameworks are measured in different ways.
\codebase~reports the maximum memory of all GPUs, maskrcnn-benchmark reports
the memory of GPU 0, and these two adopt the PyTorch API
``torch.cuda.max\_memory\_allocated()''.
Detectron reports the GPU with the caffe2 API
``caffe2.python.utils.GetGPUMemoryUsageStats()'', and SimpleDet reports the
memory shown by ``nvidia-smi'', a command line utility provided by NVIDIA.
Generally, the actual memory usage of \codebase~and maskrcnn-benchmark are
similar and lower than the others.

\begin{table*}
    \centering
	\caption{Comparison of different codebases in terms of speed, memory and performance.}
    \label{tab:codebase-compare}
    \begin{tabular}{c|cccccc}
		\hline
		Codebase            & model & Train (iter/s) & Inf (fps) & Mem (GB) & AP$_{box}$ & AP$_{mask}$ \\
		\hline
		\codebase           & Mask RCNN & 0.430 & 10.8 & 3.8 & 37.4 & 34.3 \\
		maskrcnn-benchmark  & Mask RCNN & 0.436 & 12.1 & 3.3 & 37.8 & 34.2 \\
		Detectron           & Mask RCNN & 0.744 & 8.1  & 8.8 & 37.8 & 34.1 \\
		SimpleDet           & Mask RCNN & 0.646 & 8.8  & 6.7 & 37.1 & 33.7 \\
		\hline
		\codebase           & RetinaNet & 0.285 & 13.1 & 3.4 & 35.8 & - \\
		maskrcnn-benchmark  & RetinaNet & 0.275 & 11.1 & 2.7 & 36.0 & - \\
		Detectron           & RetinaNet & 0.552 & 8.3  & 6.9 & 35.4 & - \\
		SimpleDet           & RetinaNet & 0.565 & 11.6 & 5.1 & 35.6 & - \\
		\hline
    \end{tabular}
\end{table*}

\begin{figure}
	\centering
    \includegraphics[width=\linewidth]{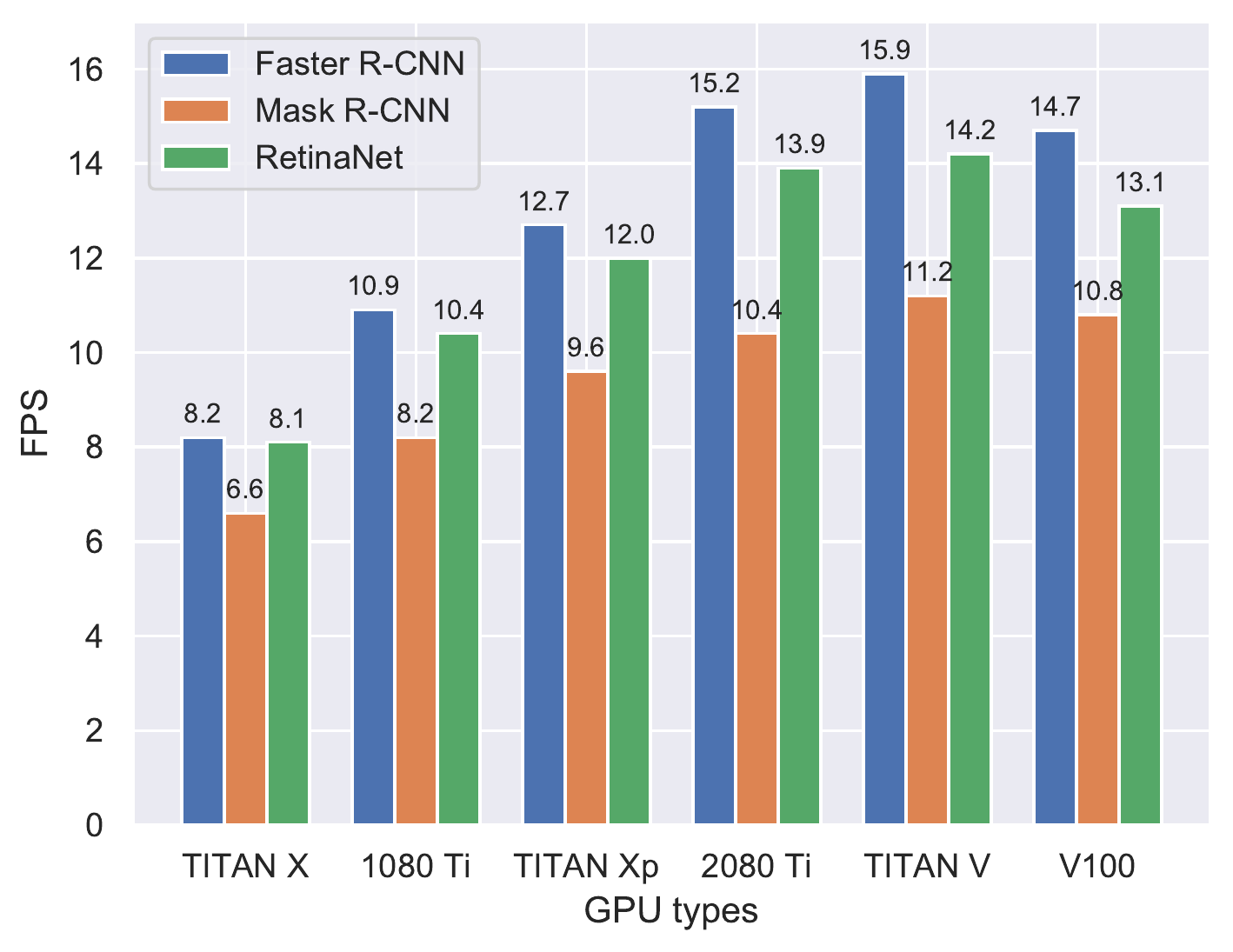}
	\caption{Inference speed benchmark of different GPUs.}
	\label{fig:speed-benchmark}
\end{figure}

\noindent\textbf{Inference speed on different GPUs.}
Different researchers may use various GPUs, here we show the speed benchmark on
common GPUs, \eg, TITAN X, TITAN Xp, TITAN V, GTX 1080 Ti, RTX 2080 Ti and V100.
We evaluate three models on each type of GPU and report the inference speed in
Figure~\ref{fig:speed-benchmark}.
It is noted that other hardwares of these servers are not exactly the same,
such as CPUs and hard disks, but the results can provide a basic impression for
the speed benchmark.

\noindent
\textbf{Mixed precision training.}
\codebase~supports mixed precision training to reduce GPU memory and to
speed up the training, while the performance remains almost the same.
The maskrcnn-benchmark supports mixed precision training with
apex\footnote{\url{https://github.com/NVIDIA/apex}} and SimpleDet also has its
own implementation. Detectron does not support it yet.
We report the results and compare with the other two codebases in Table~\ref{tab:mpt-comparison}.
We test all codebases on the same V100 node.
Additionally, we investigate more models to figure out the effectiveness of
mixed precision training.
As shown in Table~\ref{tab:mpt-results}, we can learn that a larger batch size
is more memory saving. When the batch size is increased to 12, the memory of
FP16 training is reduced to nearly half of FP32 training.
Moreover, mixed precision training is more memory efficient when applied to
simpler frameworks like RetinaNet.

\begin{table*}
    \centering
	\caption{Comparison of mixed precision training results.}
	\label{tab:mpt-comparison}
    \begin{tabular}{c|cccccc}
		\hline
		Codebase            & Type & Mem (GB) & Train (iter/s) & Inf (fps) & AP$_{box}$ & AP$_{mask}$ \\
		\hline
		\multirow{2}{*}{\codebase}
		& FP32 & 3.8      & 0.430          & 10.8       & 37.4       & 34.3 \\
		& FP16 & 3.0      & 0.364          & 10.9      & 37.4       & 34.4 \\
		\hline
		\multirow{2}{*}{maskrcnn-benchmark}
		& FP32 & 3.3      & 0.436          & 12.1      & 37.8       & 34.2 \\
		& FP16 & 3.3      & 0.457          & 9.0       & 37.7       & 34.2 \\
		\hline
		\multirow{2}{*}{SimpleDet}
		& FP32 & 6.7      & 0.646          & 8.8       & 37.1       & 33.7 \\
		& FP16 & 5.5      & 0.635          & 9.0       & 37.3       & 33.9 \\
		\hline
    \end{tabular}
\end{table*}

\begin{table}
    \centering
	\caption{Mixed precision training results of \codebase on different models. ``BS'' denotes the images of each GPU. The training memory is measured by GB and training speed is measured by s/iter.}
	\label{tab:mpt-results}
	\addtolength{\tabcolsep}{-1pt}
    \begin{tabular}{ccccccc}
		\hline
		Model                         & Backbone    & BS & Type & Mem & Speed \\
		\hline
		\multirow{8}{*}{Faster R-CNN} & R-18        & 2  & FP32 & 2.0 & 0.279  \\
		                              & R-18        & 2  & FP16 & 1.7 & 0.248  \\
		                              & R-18        & 4  & FP32 & 3.6 & 0.459  \\
		                              & R-18        & 4  & FP16 & 2.2 & 0.375  \\
		                              & R-18        & 8  & FP32 & 6.9 & 0.857  \\
		                              & R-18        & 8  & FP16 & 3.9 & 0.741  \\
		                              & R-18        & 12 & FP32 & 11.3& 1.308  \\
		                              & R-18        & 12 & FP16 & 5.7 & 1.071  \\
		\hline
		\multirow{2}{*}{Mask R-CNN}   & R-50        & 2  & FP32 & 3.8 & 0.430  \\
		                              & R-50        & 2  & FP16 & 3.0 & 0.364  \\
		\hline
		\multirow{2}{*}{RetinaNet}    & R-50        & 2  & FP32 & 3.6 & 0.308  \\
					                  & R-50        & 2  & FP16 & 2.9 & 0.232  \\
		\hline
		\multirow{2}{*}{FCOS}         & R-50        & 4  & FP32 & 6.9 & 0.396  \\
		                              & R-50        & 4  & FP16 & 5.2 & 0.270  \\
		\hline
    \end{tabular}
\end{table}

\begin{figure}
    \centering
    \includegraphics[width=\linewidth]{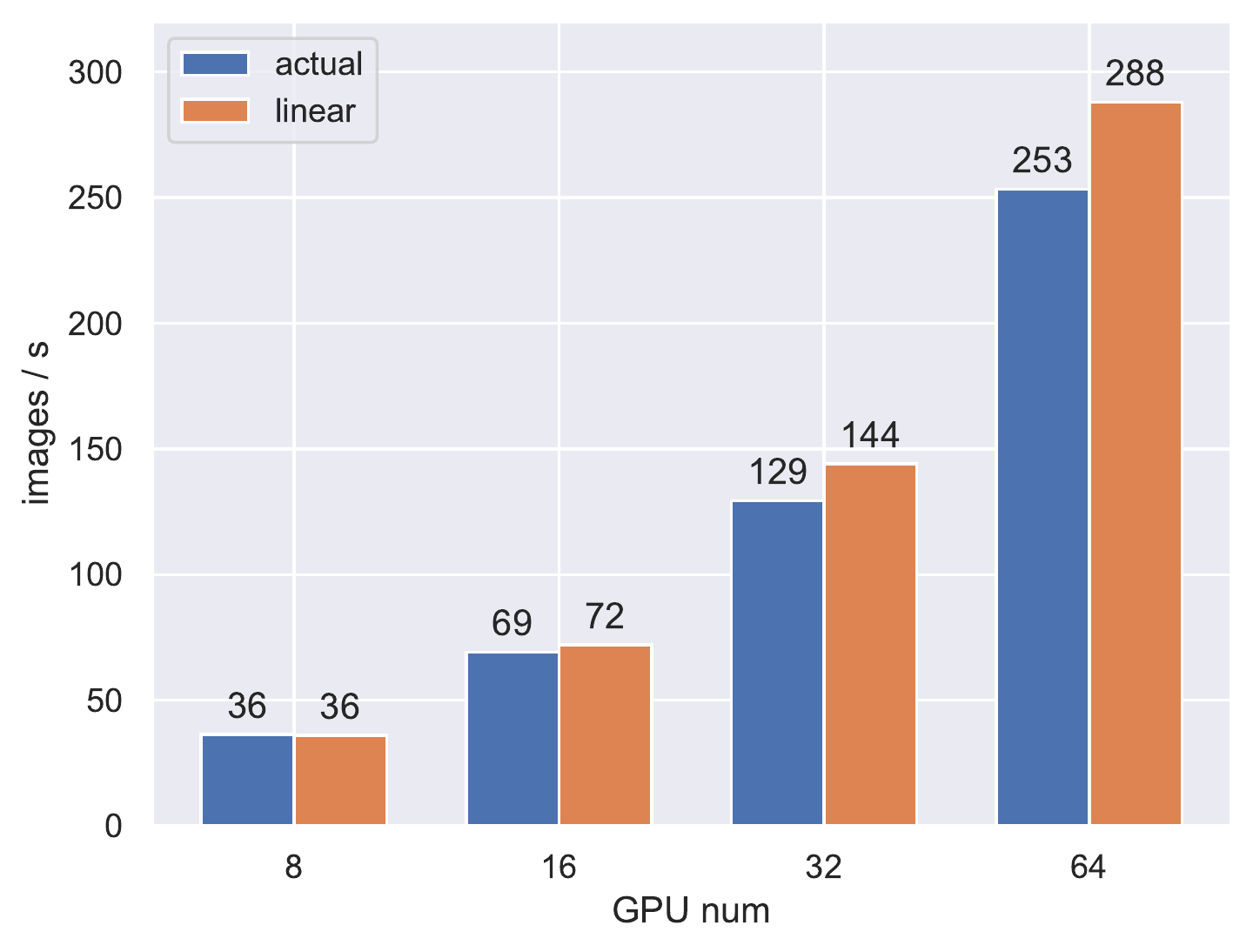}
	\caption{Training speed of Mask R-CNN on multiple nodes. The blue bar shows the performance of \codebase~and the yellow bar indicates linear speedup upper bound.}
	\vspace{-0.3cm}
    \label{fig:scalability}
\end{figure}

\noindent
\textbf{Multi-node scalability.}
Since \codebase~supports distributed training on multiple nodes, we test its
scalability on 8, 16, 32, 64 GPUs, respectively.
We adopt Mask R-CNN as the benchmarking method and conduct experiments on
another V100 cluster.
Following~\cite{goyal2017accurate}, the base learning rate is adjusted linearly
when adopting different batch sizes.
Experimental results in Figure~\ref{fig:scalability} shows that
\codebase~achieves nearly linear acceleration for multiple nodes.

\section{Extensive Studies}

With \codebase, we conducted extensive study on some important components and hyper-parameters.
We wish that the study can shed lights to better practices in making fair comparisons across different methods and settings.

\subsection{Regression Losses}
\label{subsec:regression-losses}

A multi-task loss is usually adopted for training an object detector,
which consists of the classification and regression branch.
The most widely adopted regression loss is Smooth L1 loss.
Recently, there are more regression losses proposed, \eg,
Bounded IoU Loss~\cite{tychsen2018improving}, IoU Loss~\cite{tian2019fcos},
GIoU Loss~\cite{rezatofighi2019generalized}, Balanced L1 Loss~\cite{pang2019libra}.
L1 Loss is also a straight-forward variant.
However, these losses are usually implemented in different methods and settings.
Here we evaluate all the losses under the same environment.
It is noted that the final performance varies with different loss weights
assigned to the regression loss, hence, we perform coarse grid search to find
the best loss weight for each loss.

Results in Table~\ref{tab:reg-loss} show that by simply increasing the loss
weight of Smooth L1 Loss, the final performance can improve by $0.5\%$.
Without tuning the loss weight, L1 Loss is $0.6\%$ higher than Smooth L1,
while increasing the loss weight will not bring further gain.
L1 loss has larger loss values than Smooth L1, especially for bounding boxes
that are relatively accurate.
According to the analysis in~\cite{pang2019libra}, boosting the gradients of
better located bounding boxes will benefit the localization.
The loss values of L1 loss are already quite large, therefore, increasing loss
weight does not work better.
Balanced L1 Loss achieves 0.3\% higher mAP than L1 Loss for end-to-end Faster
R-CNN, which is a little different from experiments in~\cite{pang2019libra}
that adopts pre-computed proposals. However, we find that Balanced L1 loss can
lead to a higher gain on the baseline of the proposed IoU-balanced sampling or
balanced FPN.
IoU-based losses perform slightly better than L1-based losses with optimal loss
weights except for Bounded IoU Loss. GIoU Loss is $0.1\%$ higher than IoU Loss,
and Bounded IoU Loss has similar performance to Smooth L1 Loss, but requires
a larger loss weight.

\begin{table}
    \centering
	\caption{Comparison of various regression losses with different loss
	weights (lw). Faster RCNN with ResNet-50-FPN is adopted.}
	\label{tab:reg-loss}
    \begin{tabular}{c|cccc}
		\hline
		Regression Loss & lw=1 & lw=2 & lw=5 & lw=10 \\
		\hline
		Smooth L1 Loss\cite{ren2015faster}          & $36.4$ & $36.9$ & $35.7$ & - \\
		L1 Loss                                     & $36.8$ & $36.9$ & $34.0$ & - \\
		Balanced L1 Loss\cite{pang2019libra}        & $37.2$ & $36.7$ & $33.0$ & - \\
		IoU Loss\cite{tian2019fcos}                 & $36.9$ & $37.3$ & $35.4$ & $30.7$ \\
		GIoU Loss\cite{rezatofighi2019generalized}  & $37.1$ & $37.4$ & $35.4$ & $30.0$ \\
		Bounded IoU Loss\cite{tychsen2018improving} & $34.0$ & $35.7$ & $36.8$ & $36.8$ \\
		\hline
    \end{tabular}
\end{table}

\subsection{Normalization Layers}

The batch size used when training detectors is usually small (1 or 2) due to
limited GPU memory, and thus BN layers are usually frozen as a typical convention.
There are two options for configuring BN layers.
(1) whether to update the statistics $\text{E}(x)$ and $\text{Var}(x)$, and
(2) whether to optimize affine weights $\gamma$ and $\beta$.
Following the argument names of PyTorch, we denote (1) and (2) as \emph{eval}
and \emph{requires\_grad}. $\emph{eval}=True$ means statistics are not updated,
and $\emph{requires\_grad}=True$ means $\gamma$ and $\beta$ are also optimized
during training.
Apart from freezing BN layers, there are also other normalization layers which
tackles the problem of small batch size, such as Synchronized BN
(SyncBN)~\cite{Peng2018megdet} and Group Normalization (GN)~\cite{wu2018group}.
We first evaluate different settings for BN layers in backbones, and then
compare BN with SyncBN and GN.

\noindent
\textbf{BN settings.}
We evaluate different combinations of \emph{eval} and \emph{requires\_grad} on
Mask R-CNN, under 1x and 2x training schedules.
Results in Table~\ref{tab:norm-settings} show that updating statistics with
a small batch size severely harms the performance, when we recompute statistics
(\emph{eval} is false) and fix the affine weights (\emph{requires\_grad} is false), respectively.
Compared with $\emph{eval}=True,\emph{requires\_grad}=True$, it is $3.1\%$ lower
in terms of bbox AP and $3.0\%$ lower in terms of mask AP.
Under 1x learning rate (lr) schedule, fixing the affine weights or not only
makes slightly differences, \ie, $0.1\%$. When a longer lr schedule is adopted,
making affine weights trainable outperforms fixing these weights by about $0.5\%$.
In \codebase, $\emph{eval}=True,\emph{requires\_grad}=True$ is adopted as
the default setting.

\begin{table}
	\centering
	\caption{Comparison of different BN settings and lr schedules.
	Mask RCNN with ResNet-50-FPN is adopted.}
	\label{tab:norm-settings}
	\begin{tabular}{ccc|cc}
		\hline
		\emph{eval} & \emph{requires\_grad} & lr schedule & AP$_{box}$ & AP$_{mask}$\\
		\hline
		False       & True                  & 1x          & $34.2$ & $31.2$     \\
		True        & False                 & 1x          & $37.4$ & $34.3$     \\
		True        & True                  & 1x          & $37.3$ & $34.2$     \\
		\hline
		True        & False                 & 2x          & $37.9$ & $34.6$     \\
		True        & True                  & 2x          & $38.5$ & $35.1$     \\
		\hline
	\end{tabular}
\end{table}

\noindent
\textbf{Different normalization layers.}
Batch Normalization (BN) is widely adopted in modern CNNs. However, it heavily
depends on the large batch size to precisely estimate the statistics
$\text{E}(x)$ and $\text{Var}(x)$. In object detection, the batch size is
usually much smaller than in classification, and the typical solution is to
use the statistics of pretrained backbones and not to update them during
training, denoted as FrozenBN.
More recently, SyncBN and GN are proposed and have proved their
effectiveness~\cite{wu2018group,Peng2018megdet}.
SyncBN computes mean and variance across multi-GPUs and GN divides channels of
features into groups and computes mean and variance within each group, which
help to combat against the issue of small batch sizes.
FrozenBN, SyncBN and GN can be specified in \codebase~with only simple
modifications in config files.

Here we study two questions.
(1) \emph{How do different normalization layers compare with each other?}
(2) \emph{Where to add normalization layers to detectors?}
To answer these two questions, we run three experiments of Mask R-CNN with
ResNet-50-FPN and replace the BN layers in backbones with FrozenBN, SyncBN and
GN, respectively. Group number is set to $32$ following~\cite{wu2018group}.
Other settings and model architectures are kept the same.
In~\cite{wu2018group}, the \emph{2fc} bbox head is replaced with
\emph{4conv1fc} and GN layers are also added to FPN and bbox/mask heads.
We perform another two sets of experiments to study these two changes.
Furthermore, we explore different number of convolution layers for bbox head.

Results in Table~\ref{tab:norm-layer} show that
(1) FrozenBN, SyncBN and GN achieve similar performance if we just replace BN
layers in backbones with corresponding ones.
(2) Adding SyncBN or GN to FPN and bbox/mask head will not bring further gain.
(3) Replacing the \emph{2fc} bbox head with \emph{4conv1fc} as well as adding
normalization layers to FPN and bbox/mask head improves the performance by
around $1.5\%$.
(4) More convolution layers in bbox head will lead to higher performance.

\begin{table}
    \centering
	\caption{Comparison of adopting different normalization layers and adding normalization layers on different components. (SBN is short for SyncBN.)}
	\label{tab:norm-layer}
	\addtolength{\tabcolsep}{-1pt}
    \begin{tabular}{ccc|cc}
        \hline
		Backbone & FPN & Head                 & AP$_{box}$ & AP$_{mask}$\\
		\hline
		FrozenBN & -   & - (\emph{2fc})       & $37.3$     & $34.2$     \\
		FrozenBN & -   & - (\emph{4conv1fc})  & $37.8$     & $34.2$     \\
		\hline
		SBN      & -   & -   (\emph{2fc})     & $37.4$     & $34.1$     \\
		SBN      & SBN & SBN (\emph{2fc})     & $37.4$     & $34.6$     \\
		SBN      & SBN & SBN (\emph{4conv1fc})& $38.9$     & $35.2$     \\
		\hline
		GN       & -   & -  (\emph{2fc})      & $37.4$     & $34.3$     \\
		GN       & GN  & GN (\emph{2fc})      & $37.4$     & $34.5$     \\
		GN       & GN  & GN (\emph{2conv1fc}) & $38.2$     & $35.1$     \\
		GN       & GN  & GN (\emph{4conv1fc}) & $38.8$     & $35.2$     \\
		GN       & GN  & GN (\emph{6conv1fc}) & $39.0$     & $35.4$     \\
		\hline
    \end{tabular}
\end{table}

\subsection{Training Scales}
As a typical convention, training images are resized to a predefined scale
without changing the aspect ratio. Previous studies typically prefer a scale of
$1000\times600$, and now $1333\times800$ is typically adopted.
In \codebase, we adopt $1333\times800$ as the default training scale.
As a simple data augmentation method, multi-scale training is also commonly used.
No systematic study exists to examine the way to select an appropriate training
scales. Knowing this is crucial to facilitate more effective and efficient training.
When multi-scale training is adopted, a scale is randomly selected in each
iteration, and the image will be resized to the selected scale.
There are mainly two random selection methods, one is to predefine a set of
scales and randomly pick a scale from them, the other is to define a scale
range, and randomly generate a scale between the minimum and maximum scale.
We denote the first method as ``value'' mode and the second one as ``range''
mode. Specifically, ``range'' mode can be seen as a special case of ``value''
mode where the interval of predefined scales is 1.

We train Mask R-CNN with different scales and random modes, and adopt the 2x
lr schedule because more training augmentation usually requires longer lr
schedules. The results are shown in Table~\ref{tab:training-scale},
in which $\text{1333}\times\text{[640:800:32]}$ indicates that the longer edge
is fixed to 1333 and the shorter edge is randomly selected from the pool of
$\{640, 672, 704, 736, 768, 800\}$, corresponding to the ``value'' mode.
The setting $\text{1333}\times\text{[640:800]}$ indicates that the shorter edge
is randomly selected between $640$ and $800$, which corresponds to the ``range'' mode.
From the results we can learn that the ``range'' mode performs similar to or
slightly better than the ``value'' mode with the same minimum and maximum scales.
Usually a wider range brings more improvement, especially for larger maximum
scales. Specifically, $[640:960]$ is $0.4\%$ and $0.5\%$ higher than
$[640:800]$ in terms of bbox and mask AP.
However, a smaller minimum scale like $480$ will not achieve better performance.

\begin{table}
    \centering
	\caption{Comparison of different training scales. Mask RCNN with
	ResNet-50-FPN and 2x lr schedule are adopted.}
    \label{tab:training-scale}
    \begin{tabular}{c|cc}
        \hline
		Training scale(s)        & AP$_{box}$ & AP$_{mask}$\\
		\hline
		$1333\times800$          & $38.5$     & $35.1$     \\
		\hline
		$1333\times[640:800:32]$ & $39.3$     & $35.8$     \\
		$1333\times[640:960:32]$ & $39.7$     & $36.0$     \\
		$2000\times[640:800:32]$ & $39.3$     & $35.9$     \\
		\hline
		$1333\times[640:800]$    & $39.3$     & $35.9$     \\
		$1333\times[640:960]$    & $39.7$     & $36.3$     \\
		$1333\times[480:960]$    & $39.7$     & $36.1$     \\
		\hline
    \end{tabular}
\end{table}

\subsection{Other Hyper-parameters.}
\codebase~mainly follows the hyper-parameter settings in Detectron and also
explores our own implementations.
Empirically, we found that some of the hyper-parameters of Detectron are not
optimal, especially for RPN.
In Table~\ref{tab:hyper-parameters}, we list those that can further improve the
performance of RPN.
Although the tuning may benefit the performance,  in \codebase~we adopt the
same setting as Detectron by default and just leave this study for reference.

\begin{table}
	\centering
	\caption{Study of hyper-parameters on RPN ResNet-50.}
	\label{tab:hyper-parameters}
	\begin{tabular}{c|c|c|c}
		\hline
		smoothl1\_beta & allowed\_border & neg\_pos\_ub   & AR$_{1000}$ \\
		\hline
			$1/5$     &  0         & $\infty$       & 56.5     \\
	    	$1/9$     &  0         & $\infty$       & 57.1      \\
			$1/15$    &  0         & $\infty$       & 57.3      \\
			$1/9$     &  $\infty$  & $\infty$       & 57.7      \\
			$1/9$     &  $\infty$  & 3              & 58.3      \\
			$1/9$     &  $\infty$  & 5              & 58.1      \\
		\hline
	\end{tabular}
\end{table}

\noindent\textbf{smoothl1\_beta}
Most detection methods adopt Smooth L1 Loss as the regression loss,
implemented as $torch.where(x < beta, 0.5 * x^2 / beta, x - 0.5 * beta)$.
The parameter $beta$ is the threshold for L1 term and MSELoss term.
It is set to $\frac{1}{9}$ in RPN by default, according to the
standard deviation of regression errors empirically.
Experimental results show that a smaller $beta$ may improve average recall (AR)
of RPN slightly.
In the study of Section~\ref{subsec:regression-losses}, we found that L1
Loss performs better than Smooth L1 when the loss weight is 1.
When we set $beta$ to a smaller value, Smooth L1 Loss will get closer to L1
Loss and the equivalent loss weight is larger, resulting in better performance.

\noindent\textbf{allowed\_border}
In RPN, pre-defined anchors are generated on each location of a feature map.
Anchors exceeding the boundaries of the image by more than
$\text{allowed\_border}$ will be ignored during training.
It is set to 0 by default, which means any anchors exceeding the image boundary
will be ignored. However, we find that relaxing this rule will be beneficial.
If we set it to infinity, which means none of the anchors are ignored, AR will
be improved from $57.1\%$ to $57.7\%$.
In this way, ground truth objects near boundaries will have more matching
positive samples during training.

\noindent\textbf{neg\_pos\_ub}
We add this new hyper-parameter for sampling positive and negative anchors.
When training the RPN, in the case when insufficient positive anchors are present,
one typically samples more negative samples to guarantee a fixed number of training samples.
Here we explore $\text{neg\_pos\_ub}$ to control the upper bound of the ratio
of negative samples to positive samples.
Setting $\text{neg\_pos\_ub}$ to infinity leads to the aforementioned sampling behavior.
This default practice will sometimes cause imbalance distribution in negative and positive samples.
By setting it to a reasonable value, \eg, 3 or 5, which means we sample
negative samples at most 3 or 5 times of positive ones, a gain of $1.2\%$ or
$1.1\%$ is observed.
\appendix
\section{Detailed Results}

We present detailed benchmarking results for some methods in Table~\ref{tab:detailed-results}.
R-50 and R-50 (c) denote pytorch-style and caffe-style ResNet-50 backbone,
respectively. In the bottleneck residual block, pytorch-style ResNet uses a 1x1
stride-1 convolutional layer followed by a 3x3 stride-2 convolutional layer,
while caffe-style ResNet uses a 1x1 stride-2 convolutional layer followed by
a 3x3 stride-1 convolutional layer.
Refer to \url{https://github.com/open-mmlab/mmdetection/blob/master/MODEL_ZOO.md} for more settings and components.

\onecolumn
\begin{small}
\addtolength{\tabcolsep}{-2pt}
\begin{longtable}{l|l|c|cccccc|cccccc}
	\caption{Results of different detection methods on COCO \emph{val2017}. $\text{AP}^{b}$ and $\text{AP}^{m}$ denote box mAP and mask mAP respectively.}
	\label{tab:detailed-results}\\
	\hline
	Method & Backbone & Lr Schd & $\text{AP}^{b}$ & $\text{AP}^{b}_{50}$ & $\text{AP}^{b}_{75}$ & $\text{AP}^{b}_{S}$ & $\text{AP}^{b}_{M}$ & $\text{AP}^{b}_{L}$ & $\text{AP}^{m}$ & $\text{AP}^{m}_{50}$ & $\text{AP}^{m}_{75}$ & $\text{AP}^{m}_{S}$ & $\text{AP}^{m}_{M}$ & $\text{AP}^{m}_{L}$ \\
	\hline
	\endfirsthead
	\multicolumn{15}{c}%
	{\tablename\ \thetable\ -- \textit{Continued from previous page}} \\
	\hline
	Method & Backbone & Lr Schd & $\text{AP}^{b}$ & $\text{AP}^{b}_{50}$ & $\text{AP}^{b}_{75}$ & $\text{AP}^{b}_{S}$ & $\text{AP}^{b}_{M}$ & $\text{AP}^{b}_{L}$ & $\text{AP}^{m}$ & $\text{AP}^{m}_{50}$ & $\text{AP}^{m}_{75}$ & $\text{AP}^{m}_{S}$ & $\text{AP}^{m}_{M}$ & $\text{AP}^{m}_{L}$ \\
	\hline
	\endhead
	\hline \multicolumn{15}{r}{\textit{Continued on next page}} \\
	\endfoot
	\hline
	\endlastfoot
	\multirow{10}{*}{Faster R-CNN}
	& R-50 (c)    & 1x & 36.6 & 58.5 & 39.2 & 20.7 & 40.5 & 47.9 & -    & -    & -    & -    & -    & -   \\
	& R-101 (c)   & 1x & 38.8 & 60.5 & 42.3 & 23.3 & 43.1 & 50.3 & -    & -    & -    & -    & -    & -   \\
	& R-50        & 1x & 36.4 & 58.4 & 39.1 & 21.5 & 40.0 & 46.6 & -    & -    & -    & -    & -    & -   \\
	& R-101       & 1x & 38.5 & 60.3 & 41.6 & 22.3 & 43.0 & 49.8 & -    & -    & -    & -    & -    & -   \\
	& X-101-32x4d & 1x & 40.1 & 62.0 & 43.8 & 23.4 & 44.6 & 51.7 & -    & -    & -    & -    & -    & -   \\
	& X-101-64x4d & 1x & 41.3 & 63.3 & 45.2 & 24.4 & 45.8 & 53.4 & -    & -    & -    & -    & -    & -   \\
	& R-50        & 2x & 37.7 & 59.2 & 41.1 & 21.9 & 41.4 & 48.7 & -    & -    & -    & -    & -    & -   \\
	& R-101       & 2x & 39.4 & 60.6 & 43.0 & 22.1 & 43.6 & 52.1 & -    & -    & -    & -    & -    & -   \\
	& X-101-32x4d & 2x & 40.4 & 61.9 & 44.1 & 23.3 & 44.6 & 52.9 & -    & -    & -    & -    & -    & -   \\
	& X-101-64x4d & 2x & 40.7 & 62.0 & 44.6 & 22.9 & 44.5 & 53.6 & -    & -    & -    & -    & -    & -   \\
	\hline
	\multirow{8}{*}{Cascade R-CNN}
	& R-50        & 1x & 40.4 & 58.5 & 43.9 & 21.5 & 43.7 & 53.8 & -    & -    & -    & -    & -    & -   \\
	& R-101       & 1x & 42.0 & 60.3 & 45.9 & 23.2 & 45.9 & 56.3 & -    & -    & -    & -    & -    & -   \\
	& X-101-32x4d & 1x & 43.6 & 62.2 & 47.4 & 25.0 & 47.7 & 57.4 & -    & -    & -    & -    & -    & -   \\
	& X-101-64x4d & 1x & 44.5 & 63.3 & 48.6 & 26.1 & 48.1 & 59.1 & -    & -    & -    & -    & -    & -   \\
	& R-50        & 20e& 41.1 & 59.1 & 44.8 & 22.5 & 44.4 & 54.9 & -    & -    & -    & -    & -    & -   \\
	& R-101       & 20e& 42.5 & 60.7 & 46.3 & 23.7 & 46.1 & 56.9 & -    & -    & -    & -    & -    & -   \\
	& X-101-32x4d & 20e& 44.0 & 62.5 & 48.0 & 25.3 & 47.8 & 58.1 & -    & -    & -    & -    & -    & -   \\
	& X-101-64x4d & 20e& 44.7 & 63.1 & 49.0 & 25.8 & 48.3 & 58.8 & -    & -    & -    & -    & -    & -   \\
	\hline
	SSD300 & VGG16 & 120e & 25.7 & 43.9 & 26.2 & 6.9  & 27.7 & 42.6 & -    & -    & -    & -    & -    & -   \\
	SSD512 & VGG16 & 120e & 29.3 & 49.2 & 30.8 & 11.8 & 34.1 & 44.7 & -    & -    & -    & -    & -    & -   \\
	\hline
	\multirow{10}{*}{RetinaNet}
	& R-50 (c)    & 1x & 35.8 & 55.5 & 38.3 & 20.1 & 39.5 & 47.7 & -    & -    & -    & -    & -    & -   \\
	& R-101 (c)   & 1x & 37.8 & 58.0 & 40.7 & 20.4 & 42.1 & 50.7 & -    & -    & -    & -    & -    & -   \\
	& R-50        & 1x & 35.6 & 55.5 & 38.3 & 20.0 & 39.6 & 46.8 & -    & -    & -    & -    & -    & -   \\
	& R-101       & 1x & 37.7 & 57.5 & 40.4 & 21.1 & 42.2 & 49.5 & -    & -    & -    & -    & -    & -   \\
	& X-101-32x4d & 1x & 39.0 & 59.4 & 41.7 & 22.6 & 43.4 & 50.9 & -    & -    & -    & -    & -    & -   \\
	& X-101-64x4d & 1x & 40.0 & 60.9 & 43.0 & 23.5 & 44.4 & 52.6 & -    & -    & -    & -    & -    & -   \\
	& R-50        & 2x & 36.4 & 56.3 & 38.7 & 19.3 & 39.9 & 48.9 & -    & -    & -    & -    & -    & -   \\
	& R-101       & 2x & 38.1 & 58.1 & 40.6 & 20.2 & 41.8 & 50.8 & -    & -    & -    & -    & -    & -   \\
	& X-101-32x4d & 2x & 39.3 & 59.8 & 42.3 & 21.0 & 43.6 & 52.3 & -    & -    & -    & -    & -    & -   \\
	& X-101-64x4d & 2x & 39.6 & 60.3 & 42.3 & 21.6 & 43.5 & 53.5 & -    & -    & -    & -    & -    & -   \\
	\hline
	\multirow{4}{*}{RetinaNet-GHM}
	& R-50        & 1x & 36.9 & 55.5 & 39.1 & 20.4 & 40.3 & 48.7 & -    & -    & -    & -    & -    & -   \\
	& R-101       & 1x & 39.0 & 57.7 & 41.3 & 21.8 & 43.2 & 51.8 & -    & -    & -    & -    & -    & -   \\
	& X-101-32x4d & 1x & 40.5 & 59.7 & 43.1 & 22.8 & 44.8 & 53.5 & -    & -    & -    & -    & -    & -   \\
	& X-101-64x4d & 1x & 41.6 & 61.3 & 44.3 & 23.5 & 45.5 & 55.1 & -    & -    & -    & -    & -    & -   \\
	\hline
	\multirow{4}{*}{FCOS}
	& R-50 (c)    & 1x & 36.7 & 55.8 & 39.2 & 21.0 & 40.7 & 48.4 & -    & -    & -    & -    & -    & -   \\
	& R-101 (c)   & 1x & 39.1 & 58.5 & 41.8 & 22.0 & 43.5 & 51.1 & -    & -    & -    & -    & -    & -   \\
	& R-50 (c)    & 2x & 36.9 & 55.8 & 39.1 & 20.4 & 40.1 & 49.2 & -    & -    & -    & -    & -    & -    \\
	& R-101 (c)   & 2x & 39.1 & 58.6 & 41.7 & 22.1 & 42.4 & 52.5 & -    & -    & -    & -    & -    & -    \\
	\hline
	\multirow{3}{*}{FCOS (mstrain)}
	& R-50 (c)    & 2x & 38.7 & 58.0 & 41.4 & 23.4 & 42.8 & 49.0 & -    & -    & -    & -    & -    & -   \\
	& R-101 (c)   & 2x & 40.8 & 60.1 & 43.8 & 24.5 & 44.5 & 52.8 & -    & -    & -    & -    & -    & -   \\
	& X-101-64x4d & 2x & 42.8 & 62.6 & 45.7 & 26.5 & 46.9 & 54.5 & -    & -    & -    & -    & -    & -   \\
	\hline
	\multirow{4}{*}{Libra Faster R-CNN}
	& R-50        & 1x & 38.5 & 59.5 & 42.5 & 22.9 & 41.8 & 48.9 & -    & -    & -    & -    & -    & -   \\
	& R-101       & 1x & 40.3 & 61.2 & 43.9 & 23.3 & 44.3 & 52.2 & -    & -    & -    & -    & -    & -   \\
	& X-101-32x4d & 1x & 41.6 & 62.7 & 45.6 & 24.8 & 45.8 & 53.6 & -    & -    & -    & -    & -    & -   \\
	& X-101-64x4d & 1x & 42.7 & 63.8 & 46.8 & 25.8 & 46.6 & 55.4 & -    & -    & -    & -    & -    & -   \\
	\hline
	\multirow{4}{*}{GA-Faster R-CNN}
	& R-50 (c)    & 1x & 39.9 & 59.1 & 43.6 & 22.8 & 43.5 & 52.8 & -    & -    & -    & -    & -    & -   \\
	& R-101 (c)   & 1x & 41.5 & 60.7 & 45.5 & 23.3 & 45.6 & 55.3 & -    & -    & -    & -    & -    & -   \\
	& X-101-32x4d & 1x & 42.9 & 62.1 & 46.8 & 24.8 & 46.9 & 56.1 & -    & -    & -    & -    & -    & -   \\
	& X-101-64x4d & 1x & 43.9 & 63.3 & 48.3 & 25.4 & 47.9 & 57.0 & -    & -    & -    & -    & -    & -   \\
	\hline
	\multirow{4}{*}{GA-RetinaNet}
	& R-50 (c)    & 1x & 37.0 & 56.6 & 39.8 & 20.0 & 40.8 & 50.1 & -    & -    & -    & -    & -    & -   \\
	& R-101 (c)   & 1x & 38.9 & 59.1 & 41.8 & 22.0 & 42.6 & 51.9 & -    & -    & -    & -    & -    & -   \\
	& X-101-32x4d & 1x & 40.3 & 60.9 & 43.5 & 23.5 & 44.9 & 53.5 & -    & -    & -    & -    & -    & -   \\
	& X-101-64x4d & 1x & 40.8 & 61.4 & 44.0 & 23.9 & 44.9 & 54.3 & -    & -    & -    & -    & -    & -   \\
	\hline
	\multirow{10}{*}{Mask R-CNN}
	& R-50 (c)    & 1x & 37.4 & 58.9 & 40.4 & 21.7 & 41.0 & 49.1 & 34.3 & 55.8 & 36.4 & 18.0 & 37.6 & 47.3 \\
	& R-101 (c)   & 1x & 39.9 & 61.5 & 43.6 & 23.9 & 44.0 & 51.8 & 36.1 & 57.9 & 38.7 & 19.8 & 39.8 & 49.5 \\
	& R-50        & 1x & 37.3 & 59.0 & 40.2 & 21.9 & 40.9 & 48.1 & 34.2 & 55.9 & 36.2 & 18.2 & 37.5 & 46.3 \\
	& R-101       & 1x & 39.4 & 60.9 & 43.3 & 23.0 & 43.7 & 51.4 & 35.9 & 57.7 & 38.4 & 19.2 & 39.7 & 49.7 \\
	& X-101-32x4d & 1x & 41.1 & 62.8 & 45.0 & 24.0 & 45.4 & 52.6 & 37.1 & 59.4 & 39.8 & 19.7 & 41.1 & 50.1 \\
	& X-101-64x4d & 1x & 42.1 & 63.8 & 46.3 & 24.4 & 46.6 & 55.3 & 38.0 & 60.6 & 40.9 & 20.2 & 42.1 & 52.4 \\
	& R-50        & 2x & 38.5 & 59.9 & 41.8 & 22.6 & 42.0 & 50.5 & 35.1 & 56.8 & 37.0 & 18.9 & 38.0 & 48.3 \\
	& R-101       & 2x & 40.3 & 61.5 & 44.1 & 22.2 & 44.8 & 52.9 & 36.5 & 58.1 & 39.1 & 18.4 & 40.2 & 50.4 \\
	& X-101-32x4d & 2x & 41.4 & 62.5 & 45.4 & 24.0 & 45.4 & 54.5 & 37.1 & 59.4 & 39.5 & 19.9 & 40.6 & 51.3 \\
	& X-101-64x4d & 2x & 42.0 & 63.1 & 46.1 & 23.9 & 45.8 & 55.6 & 37.7 & 59.9 & 40.4 & 19.6 & 41.3 & 52.5 \\
	\hline
	\multirow{5}{*}{Mask Scoring R-CNN}
	& R-50 (c)    & 1x & 37.5 & 59.2 & 40.5 & 21.4 & 41.3 & 48.9 & 35.6 & 55.6 & 38.5 & 18.2 & 39.1 & 49.2 \\
	& R-101 (c)   & 1x & 40.0 & 61.4 & 43.7 & 23.2 & 44.2 & 52.3 & 37.3 & 57.7 & 40.2 & 19.5 & 41.1 & 51.6 \\
	& X-101-64x4d & 1x & 42.2 & 64.0 & 46.2 & 24.9 & 46.5 & 54.6 & 39.2 & 60.4 & 42.4 & 21.1 & 43.1 & 54.3 \\
	& X-101-32x4d & 2x & 41.5 & 62.6 & 45.1 & 23.7 & 45.2 & 54.7 & 38.4 & 58.9 & 41.7 & 20.1 & 42.0 & 53.9 \\
	& X-101-64x4d & 2x & 42.2 & 63.4 & 46.1 & 24.2 & 46.0 & 56.1 & 38.9 & 59.4 & 42.1 & 20.4 & 42.4 & 54.7 \\
	\hline
	\multirow{8}{*}{Cascade Mask R-CNN}
	& R-50        & 1x  & 41.2 & 59.1 & 45.1 & 23.3 & 44.5 & 54.5 & 35.7 & 56.3 & 38.6 & 18.5 & 38.6 & 49.2 \\
	& R-101       & 1x  & 42.6 & 60.7 & 46.7 & 23.8 & 46.4 & 56.9 & 37.0 & 58.0 & 39.9 & 19.1 & 40.5 & 51.4 \\
	& X-101-32x4d & 1x  & 44.4 & 62.6 & 48.6 & 25.4 & 48.1 & 58.7 & 38.2 & 59.6 & 41.2 & 20.3 & 41.9 & 52.4 \\
	& X-101-64x4d & 1x  & 45.4 & 63.7 & 49.7 & 25.8 & 49.2 & 60.6 & 39.1 & 61.0 & 42.1 & 20.5 & 42.6 & 54.1 \\
	& R-50        & 20e & 42.3 & 60.5 & 46.0 & 23.7 & 45.7 & 56.4 & 36.6 & 57.6 & 39.5 & 19.0 & 39.4 & 50.7 \\
	& R-101       & 20e & 43.3 & 61.3 & 47.0 & 24.4 & 46.9 & 58.0 & 37.6 & 58.5 & 40.6 & 19.7 & 40.8 & 52.4 \\
	& X-101-32x4d & 20e & 44.7 & 63.0 & 48.9 & 25.9 & 48.7 & 58.9 & 38.6 & 60.2 & 41.7 & 20.9 & 42.1 & 52.7 \\
	& X-101-64x4d & 20e & 45.7 & 64.1 & 50.0 & 26.2 & 49.6 & 60.0 & 39.4 & 61.3 & 42.9 & 20.8 & 42.7 & 54.1 \\
	\hline
	\multirow{5}{*}{Hyrbrid Task Cascade}
	& R-50        & 1x  & 42.1 & 60.8 & 45.9 & 23.9 & 45.5 & 56.2 & 37.3 & 58.2 & 40.2 & 19.5 & 40.6 & 51.7 \\
	& R-50        & 20e & 43.2 & 62.1 & 46.8 & 24.9 & 46.4 & 57.8 & 38.1 & 59.4 & 41.0 & 20.3 & 41.1 & 52.8 \\
	& R-101       & 20e & 44.9 & 63.8 & 48.7 & 26.4 & 48.3 & 59.9 & 39.4 & 60.9 & 42.4 & 21.4 & 42.4 & 54.4 \\
	& X-101-32x4d & 20e & 46.1 & 65.1 & 50.2 & 27.5 & 49.8 & 61.2 & 40.3 & 62.2 & 43.5 & 22.3 & 43.7 & 55.5 \\
	& X-101-64x4d & 20e & 46.9 & 66.0 & 51.2 & 28.0 & 50.7 & 62.1 & 40.8 & 63.3 & 44.1 & 22.7 & 44.2 & 56.3 \\
\end{longtable}
\end{small}
\twocolumn

{\small
	\bibliographystyle{ieee_fullname}
	\bibliography{sections/egbib}
}

\end{document}